\documentclass[runningheads]{llncs}

\usepackage{amssymb}

\usepackage{booktabs}
\usepackage{multirow}
\usepackage[section]{placeins}
\usepackage{cite}
\usepackage[colorlinks,
            linkcolor=blue,
            anchorcolor=blue,
            citecolor=blue]{hyperref}
\usepackage{subfigure}
\makeatletter 
\renewcommand{\@thesubfigure}{\hskip\subfiglabelskip}
\makeatother
\usepackage{overpic}
\usepackage{caption}
\usepackage{graphicx}
\graphicspath{{figures/}}
\usepackage{marvosym}

\begin{document}
    \title{Correlation-Aware Mutual Learning for Semi-supervised Medical Image Segmentation}
    \titlerunning{Correlation-Aware Mutual Learning} 
    
    \author
    {
        Shengbo Gao    \inst{1\star} 
        \and
        Ziji Zhang     \inst{2\star\dagger} 
        \and
        Jiechao Ma     \inst{1}
        \and
        Zihao Li       \inst{1} 
        \and
        Shu Zhang       $^{1}$ \href{mailto:zhangshu@deepwise.com}{\textrm{\Letter}}
    }
    
    \authorrunning{S. Gao and Z. Zhang et al.}
    
    \institute{
    $^{1}$ Deepwise AI Lab, Beijing, China
    \\
    $^{2}$ School of Artificial Intelligence, Beijing University of Posts and Telecommunications
    \\
    \email{zhangshu@deepwise.com} 
    }

    \maketitle              
    \begin{abstract}
        Semi-supervised learning has become increasingly popular in medical image segmentation due to its ability to leverage large amounts of unlabeled data to extract additional information. 
        However, 
        most existing semi-supervised segmentation methods only focus on extracting information from unlabeled data, 
        disregarding the potential of labeled data to further improve the performance of the model.
        In this paper, 
        we propose a novel Correlation Aware Mutual Learning (CAML) framework that leverages labeled data to guide the extraction of information from unlabeled data. 
        Our approach is based on a mutual learning strategy that incorporates two modules: 
        the Cross-sample Mutual Attention Module (CMA) and the Omni-Correlation Consistency Module (OCC). 
        The CMA module establishes dense cross-sample correlations among a group of samples, 
        enabling the transfer of label prior knowledge to unlabeled data. 
        The OCC module constructs omni-correlations between the unlabeled and labeled datasets and regularizes dual models by constraining the omni-correlation matrix of each sub-model to be consistent. 
        Experiments on the Atrial Segmentation Challenge dataset demonstrate that our proposed approach outperforms state-of-the-art methods, 
        highlighting the effectiveness of our framework in medical image segmentation tasks.
        The codes, pre-trained weights, and data are publicly available. \footnote{\url{https://github.com/Herschel555/CAML}} 
        \footnote{$\star$ Both authors contributed equally to this work.} 
        \footnote{$\dagger$ Work done as an intern in Deepwise AI Lab}

    \keywords{Semi-supervised learning  \and  Medical Image Segmentation. \and Mutual learning \and Cross-sample correlation}
    \end{abstract}

    \section{Introduction}

            Despite the remarkable advancements achieved through the use of deep learning for automatic medical image segmentation, 
            the scarcity of precisely annotated training data remains a significant obstacle to the widespread adoption of such techniques in clinical settings. 
            As a solution, 
            the concept of semi-supervised segmentation has been proposed to enable models to be trained using less annotated but abundant unlabeled data.

            Recently, 
            methods that adopt the co-teaching~\cite{luo2022semi,wu2021semi,chen2021semi} or mutual learning~\cite{zhang2021robust} paradigm have emerged as a promising approach for semi-supervised learning. 
            Those methods adopt two simultaneously updated models, 
            each trained to predict the prediction results of its counterpart, 
            which can be seen as a combination of the notions of consistency regularization\cite{berthelot2019mixmatch,french2019semi,ouali2020semi,liu2022perturbed,mittal2019semi} and entropy minimization\cite{cascante2021curriculum,yuan2021simple,yang2022st++,lee2013pseudo,xie2020self}. 
            In the domain of semi-supervised medical image segmentation, 
            MC-Net~\cite{wu2021semi} has shown significant improvements in segmentation performance.  

            With the rapid advancement of semi-supervised learning, 
            the importance of unlabeled data has garnered increased attention across various disciplines in recent years. 
            However, 
            the role of labeled data has been largely overlooked, 
            with the majority of semi-supervised learning techniques treating labeled data supervision as merely an initial step of the training pipeline or as a means to ensure training convergence\cite{ouali2020semi,zou2020pseudoseg,kwon2022semi}. 
            Recently, 
            methods that can leverage labeled data to directly guide information extraction from unlabeled data have attracted the attention of the community\cite{wu2023querying}. 
            In the domain of semi-supervised medical image segmentation, there exist shared characteristics between labeled and unlabeled data that possess greater intuitiveness and instructiveness for the algorithm.
            Typically, 
            partially labeled clinical datasets exhibit similar foreground features, 
            including comparable texture, shape, and appearance among different samples.
            As such, 
            it can be hypothesized that constructing a bridge across the entire training dataset to connect labeled and unlabeled data can effectively transfer prior knowledge from labeled data to unlabeled data and facilitate the extraction of information from unlabeled data, 
            ultimately overcoming the performance bottleneck of semi-supervised learning methods.

            Based on the aforementioned conception, 
            we propose a novel \emph{Correlation Aware Mutual Learning (CAML)} framework to explicitly model the relationship between labeled and unlabeled data to effectively utilize the labeled data.
            Our proposed method incorporates two essential components, 
            namely the \emph{Cross-sample Mutual Attention module (CMA)} and the \emph{Omni-Correlation Consistency module (OCC)}, 
            to enable the effective transfer of labeled data information to unlabeled data. 
            The \emph{CMA} module establishes mutual attention among a group of samples, 
            leading to a mutually reinforced representation of co-salient features between labeled and unlabeled data. 
            Unlike conventional methods, 
            where supervised signals from labeled and unlabeled samples are separately back-propagated, 
            the proposed \emph{CMA} module creates a new information propagation path among each pixel in a group of samples, 
            which synchronously enhances the feature representation ability of each  intra-group sample. 
            
            In addition to the \emph{CMA} module, 
            we introduce the \emph{OCC} module to regularize the segmentation model by explicitly modeling the omni-correlation between unlabeled features and a group of labeled features. 
            This is achieved by constructing a memory bank to store the labeled features as a reference set of features or basis vectors. 
            In each iteration, 
            a portion of features from the memory bank is utilized to calculate the omni-correlation with unlabeled features, reflecting the similarity relationship of an unlabeled pixel with respect to a set of basis vectors of the labeled data. 
            Finally, 
            we constrain the omni-correlation matrix of each sub-model to be consistent to regularize the entire framework. 
            With the proposed omni-correlation consistency, 
            the labeled data features serve as anchor groups to guide the representation learning of the unlabeled data feature and explicitly encourage the model to learn a more unified feature distribution among unlabeled data.
            
            In summary, our contributions are threefold:
            (1)We propose a novel \emph{Correlation Aware Mutual Learning (CAML)} framework that focuses on the efficient utilization of labeled data to address the challenge of semi-supervised medical image segmentation.
            (2)We introduce the \emph{Cross-sample Mutual Attention module (CMA)} and the \emph{Omni-Correlation Consistency module (OCC)} to establish cross-sample relationships directly.
            (3)Experimental results on a benchmark dataset demonstrate significant improvements over previous SOTAs, especially when only a small number of labeled images are available.

    \section{Method}
    
        \begin{figure}[!t]
            \includegraphics[width=\textwidth]{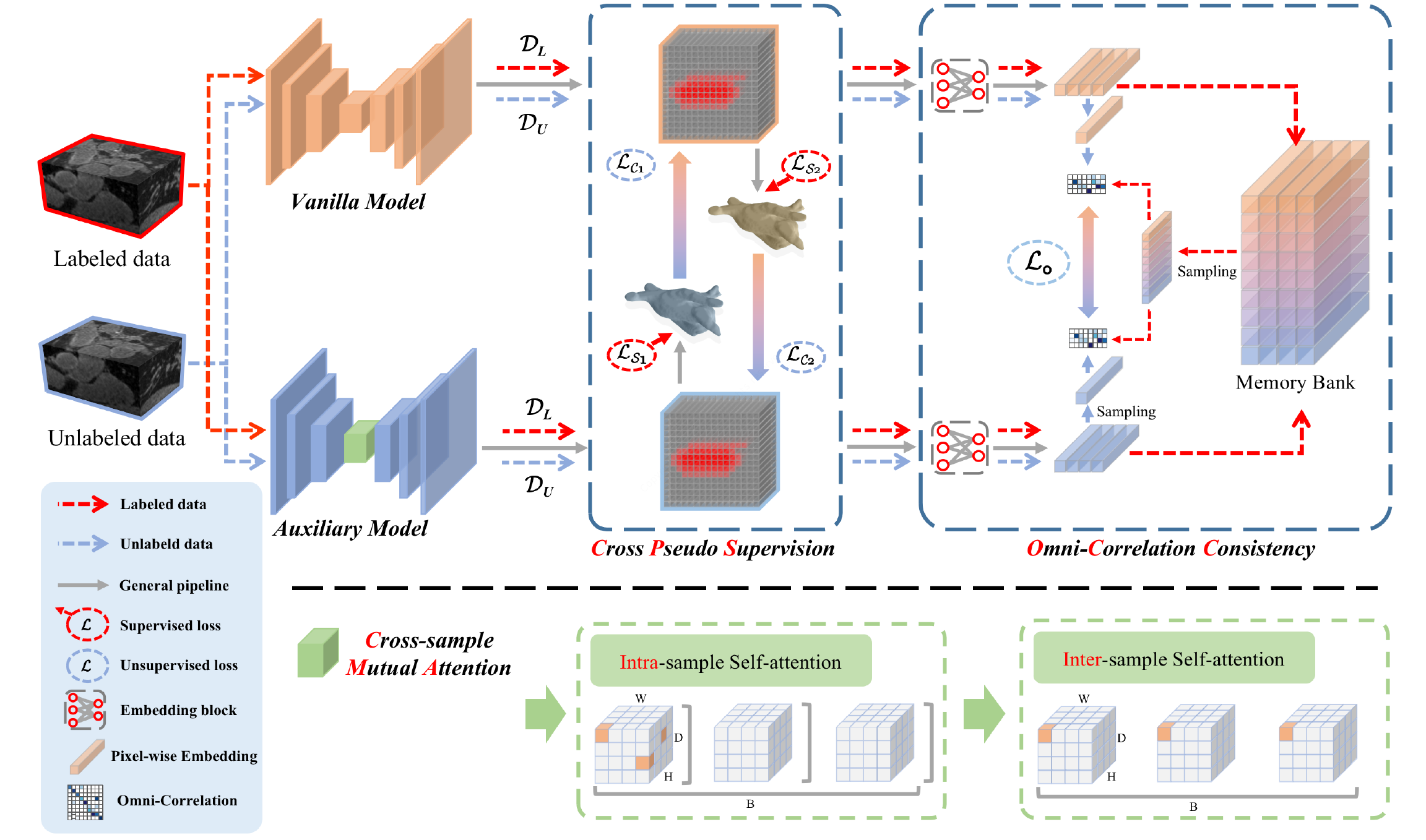}
                \caption{Overview of our proposed \emph{CAML}. \emph{CAML} adopts a co-teaching scheme with cross-pseudo supervision. The \emph{CMA} module incorporated into the auxiliary network and the \emph{OCC} module are introduced for advanced cross-sample relationship modeling.     
                } 
            \label{fig:fig1}
        \end{figure}
        
        \subsection{Overview}

            Fig \ref{fig:fig1} gives an overview of \emph{CAML}. We adopt a co-teaching paradigm like MC-Net~\cite{wu2021semi} to enforce two parallel networks to predict the prediction results of its counterpart.
            To achieve efficient cross-sample relationship modeling and enable information propagation among labeled and unlabeled data in a mini-batch,
            we incorporate a \emph{\textbf{C}ross-sample \textbf{M}utual \textbf{A}ttention} module to the auxiliary segmentation network \(f_a\),  
            whereas the vanilla segmentation network \(f_v\) remains the original V-Net structure. 
            In addition, 
            we employ an \emph{\textbf{O}mni-\textbf{C}orrelation \textbf{C}onsistency} regularization to further regularize the representation learning of the unlabeled data. Details about those two modules will be elaborated on in the following sections. The total loss of \emph{CAML} can be formulated as:
            \begin{equation}
                L=L_s+\lambda_cl_c+\lambda_ol_o
            \end{equation}
            where \(l_o\) represents the proposed omni-correlation consistency loss, while \(L_s\) and \(l_c\) are the supervised loss and the cross-supervised loss implemented in the \emph{Cross Pseudo Supervision(CPS)} module. \(\lambda_c\) and \(\lambda_o\) are the weights to control \(l_c\) and \(l_o\) separately. During the training procedure, a batch of mixed labeled and unlabeled samples are fed into the network. The supervised loss is only applied to labeled data, while all samples are utilized to construct cross-supervised learning. Please refer to~\cite{chen2021semi} for a detailed description of the \emph{CPS} module and loss design of \(L_s\) and \(l_c\).
            
        \subsection{Cross-sample Mutual Attention Module} 
            
            To enable information propagation through any positions of any samples in a mini-batch, one can simply treat each pixel's feature vector as a token and perform self-attentions for all tokens in a mini-batch. However, this will make the computation cost prohibitively large as the computation complexity of self-attention is $O(n^2)$ with respect to the number of tokens. We on the other hand adopt two sequentially mounted self-attention modules along different dimensions to enable computation efficient mutual attention among all pixels. 

            As illustrated in Fig~\ref{fig:fig1}, the proposed \emph{CMA} module consists of two sequential transformer encoder layers, 
            termed as \(E_1\) and \(E_2\), each including a multi-head attention and a \(MLP\) block with a layer normalization after each block. 
            For an input feature map \(a_{in}\in\mathbb{R}^{b\times c\times k}\), 
            where \(k=h^\prime\times w^\prime\times d^\prime\), 
            \(b\) represents batch size and \(c\) is the dimension of \(a_{in}\), 
            \(E_1\) performs intra-sample self-attention on the spatial dimension of each sample. 
            This is used to model the information propagation paths between every pixel position within each sample. 
            Then, 
            to further enable information propagation among different samples, 
            we perform an inter-sample self-attention along the batch dimension. 
            In other words, 
            along the \(b\) dimension,
            the pixels located in the same spatial position from samples are fed into a self-attention module to construct cross-sample relationships. 
            
            In \emph{CAML}, we employ the proposed \emph{CMA} module in the auxiliary segmentation network \(f_a\), 
            whereas the vanilla segmentation network \(f_v\) remains the original V-Net structure.
            The reasons can be summarized into two folds.
            From deployment perspective,
            the insertion of the CMA module requires a batch size of large than 1 to model the attention among samples within a mini-batch, 
            which is not applicable for model inference(batchsize=1). 
            From the perspective of model design,
            we model the vanilla and the auxiliary branch with different architectures to increase the architecture heterogeneous for better performance in a mutual learning framework. 
            
        \subsection{Omni-Correlation Consistency Regularization}
            In this chapter, 
            we introduce \emph{Omni-Correlation Consistency (OCC)} to formulate additional model regularization.
            The core of the \emph{OCC} module is omni-correlation,
            which is a kind of similarity matrix that is calculated between the feature of an unlabeled pixel and a group of prototype features sampled from labeled instances features. 
            It reflects the similar relationship of an unlabeled pixel with respect to a set of labeled reference pixels. 
            During the training procedure,
            we explicitly constrain the omni-correlation calculated using heterogeneous unlabeled features from those two separate branches to remain the same.
            In practice, 
            we use an Omni-correlation matrix to formulate the similarity distribution between unlabeled features and the prototype features.

            Let \(g_v\) and \(g_a\) denote two projection heads attached to the backbones of \(f_v\) and \(f_a\) separately, 
            and \(z_{v}\in\mathbb{R}^{m\times c'}\) and \(z_{a}\in\mathbb{R}^{m\times c'}\) represent two sets of embeddings sampled from their projected features extracted from unlabeled samples, 
            where \(m\) is the number of sampled features and \(c'\) is the dimension of the projected features. 
            It should be noted that \(z_v\) and \(z_a\) are sampled from the embeddings corresponding to the same set of positions on unlabeled samples. 
            Suppose \(z_{p}\in\mathbb{R}^{n\times c'}\) represents a set of prototype embeddings sampled from labeled instances, 
            where \(n\) represents the number of sampled prototype features, 
            the omni-correlation matrix calculation between \(z_v\) and \(z_p\) can be formulated as:

            \begin{equation}
                sim_{vp_{i}}=\frac{exp(cos(z_v,z_{p_{i}})*t)}{\sum_{j=1}^{n} exp(cos(z_v,z_{p_{j}})*t)}, i\in \left \{1, ..., n  \right \}  
            \end{equation}
            
            where \(cos\) means the cosine similarity and \(t\) is the temperature hyperparameter. 
            \(sim_{vp}\in\mathbb{R}^{m\times n}\) is the calculated omni-correlation matrix. Similarly, 
            the similarity distribution \(sim_{ap}\) between \(z_a\) and \(z_p\) can be calculated by replacing \(z_v\) with \(z_a\). 
            
            To constrain the consistency of omni-correlation between dual branches, the omni-correlation consistency regularization can be conducted with the cross-entropy loss \(l_{ce}\) as follows:
            
            \begin{equation}
                l_o=\frac{1}{m}\sum{l_{ce}\left(sim_{vp},sim_{ap}\right)}
            \end{equation}
            
            \noindent{\textbf{Memory Bank Construction}}
            We utilize a memory bank \(T\) to iteratively update prototype embeddings for \emph{OCC} computation.
            Specifically, 
            \(T\) initializes N slots for each labeled training sample and updates prototype embeddings with filtered labeled features projected by \(g_v\) and \(g_a\). 
            To ensure the reliability of the features stored in \(T\), 
            we select embeddings on the positions where both \(f_v\) and \(f_a\) have the correct predictions and update \(T\) with the mean fusion of the projected features projected by \(g_v\) and \(g_a\). 
            For each training sample, 
            following\cite{he2020momentum}, 
            \(T\) updates slots corresponding to the labeled samples in the current mini-batch in a query-like manner.

            \noindent{\textbf{Embeddings Sampling}}
            For computation efficiency, omni-correlation is not calculated on all labeled and unlabeled pixels. Specifically, we have developed a confidence-based mechanism to sample the pixel features from the unlabeled data.
            Practically,
            to sample \(z_v\) and \(z_a\) from unlabeled features, 
            we first select the pixels where \(f_v\) and \(f_a\) have the same prediction. 
            For each class, 
            we sort the confidence scores of these pixels, 
            and then select features of the top \(i\) pixels as the sampled unlabeled features. Thus, 
            \(m=i\times C\), 
            where \(C\) represents the number of classes.
            With regards to the prototype embeddings, we randomly sample \(j\) embeddings from each class among all the embeddings contained in \(T\) and \(n=j\times C\) to increase its diversity.

        \section{Experiments and Results}

        \noindent\textbf{Dataset.} 
    	Our method is evaluated on the Left Atrium (LA) dataset \cite{xiong2020global}
    	from the 2018 Atrial Segmentation Challenge. 
    	The dataset comprises 100 gadolinium-enhanced MR imaging scans (GE-MRIs) and their ground truth masks, with an isotropic resolution of $0.625^{3}mm^{3}$. 
    	Following \cite{yu2018pu}, 
    	we use 80 scans for training and 20 scans for testing. 
    	All scans are centered at the heart region and cropped accordingly, 
    	and then normalized to zero mean and unit variance.\\

        \noindent{\textbf{Implementation Details.}}
    	We implement our \emph{CAML} using PyTorch 1.8.1 and CUDA 10.2 on an NVIDIA TITAN RTX GPU. 
    	For training data augmentation, we randomly crop sub-volumes of size $112\times112\times80$ following \cite{yu2018pu}. 
    	To ensure a fair comparison with existing methods, 
    	we use the V-Net \cite{milletari2016v} as the backbone for all our models. 
    	During training, 
    	we use a batch size of 4,
    	with half of the images annotated and the other half unannotated. 
    	We train the entire framework using the SGD optimizer, 
    	with a learning rate of 0.01, momentum of 0.9, 
    	and weight decay of 1e-4 for 15000 iterations. 
    	To balance the loss terms in the training process, 
    	we use a time-dependent Gaussian warming up function for $\lambda_{U}$ and $\lambda_{C}$, 
    	where $\lambda(t) = \beta \ast e^{-5(1-t/t_{max})^2}$, 
    	and set $\beta$ to 1 and 0.1 for $\lambda_{U}$ and $\lambda_{C}$, 
    	respectively. 
    	For the \emph{OCC} module, 
    	we set \(c'\) to 64, 
    	j to 256, 
    	and i to 12800.
    	During inference, 
    	prediction results from the vanilla V-Net are used with a general sliding window strategy without any post-processing.

    \begin{table}[!htbp]
		\centering
		\caption{Comparison with state-of-the-art methods on the LA database.  \textbf{Metrics} reported the $mean_\pm standard$ results with three random seeds,
                         \textbf{Reported Metrics} are the results reported in the original paper. 
                }
		\resizebox{\textwidth}{!}{
			\begin{tabular}{l|cc|cccc|cccc}
				\toprule[2pt]
	      		\multicolumn{1}{c|}{\multirow{2}{*}{\textbf{Method}}}	&
				\multicolumn{2}{c|}{\textbf{Scans used}}				&
				\multicolumn{4}{c|}{\textbf{Metrics}}					&
				\multicolumn{4}{c}{\textbf{Reported Metrics}}			\\
				\cline{2-11}	&
				\multicolumn{1}{c}{Labeled}		&
				\multicolumn{1}{c|}{Unlabeled}	&
				\multicolumn{1}{c}{Dice(\%)}	&
				\multicolumn{1}{c}{Jaccard(\%)}	&
				\multicolumn{1}{c}{95HD(voxel)}	&
				\multicolumn{1}{c|}{ASD(voxel)}	&
				\multicolumn{1}{c}{Dice(\%)}	&
				\multicolumn{1}{c}{Jaccard(\%)}	&
				\multicolumn{1}{c}{95HD(voxel)}	&
				\multicolumn{1}{c|}{ASD(voxel)}	\\
				\hline
				V-Net  													& 4    							& 0								& ${43.32_{\pm8.62}}$ 					& ${31.43_{\pm6.90}}$ 					& ${40.19_{\pm1.11}}$					& ${12.13_{\pm0.57}}$   				& 52.55	& 39.60	& 47.05	& 9.87	\\
				V-Net  													& 8    							& 0								& ${79.87_{\pm1.23}}$    				& ${67.60_{\pm1.88}}$					& ${26.65_{\pm6.36}}$					& ${7.94_{\pm2.22}}$					& 78.57	& 66.96	& 21.20	& 6.07	\\
                V-Net  													& 16    						& 0								& ${85.94_{\pm0.48}}$     				& ${75.99_{\pm0.57}}$					& ${16.70_{\pm1.82}}$					& ${4.80_{\pm0.62}}$					& 86.96	& 77.31	& 11.85	& 3.22	\\
                V-Net  													& 80    						& 0								& ${90.98_{\pm0.67}}$     				& ${83.61_{\pm1.06}}$					& ${8.58_{\pm2.34}}$					& ${2.10_{\pm0.59}}$					& 91.62	& 84.60	& 5.40	& 1.64	\\
				\hline
				UA-MT	\cite{yu2018pu}						(MICCAI'19)	& \multirow{8}{*}{$4 (5\%)$}	& \multirow{8}{*}{$76 (95\%)$}	& ${78.07_{\pm0.90}}$					& $65.03_{\pm0.96}$						& $29.17_{\pm3.82}$						& $8.63_{\pm0.98}$						& - 	& -		& - 	& -		\\
				SASSNet	\cite{li2020shape}					(MICCAI'20)	& 								& 								& ${79.61_{\pm0.54}}$     				& ${67.00_{\pm0.59}}$					& ${25.54_{\pm4.60}}$					& ${7.20_{\pm1.21}}$					& - 	& - 	& - 	& -		\\
				DTC		\cite{luo2021semi}  				(AAAI'21)	& 								& 	 							& ${80.14_{\pm1.22}}$     				& ${67.88_{\pm1.82}}$					& ${24.08_{\pm2.63}}$					& ${7.18_{\pm0.62}}$					& - 	& - 	& - 	& -		\\
                MC-Net	\cite{wu2021semi}					(MedIA'21)	& 								& 	 							& ${80.92_{\pm3.88}}$     				& ${68.90_{\pm5.09}}$					& ${17.25_{\pm6.08}}$					& \textcolor{blue}{$2.76_{\pm0.49}$}	& - 	& - 	& - 	& -		\\
                URPC	\cite{media2022urpc}				(MedIA'22)	& 								& 	 							& ${80.75_{\pm0.21}}$     				& ${68.54_{\pm0.34}}$					& ${19.81_{\pm0.67}}$					& ${4.98_{\pm0.25}}$					& - 	& - 	& - 	& -		\\
                SS-Net	\cite{wu2022exploring}	            (MICCAI'22)	& 								& 	 							& \textcolor{blue}{$83.33_{\pm1.66}$}   & \textcolor{blue}{$71.79_{\pm2.36}$}	& ${15.70_{\pm0.80}}$					& ${4.33_{\pm0.36}}$					& 86.33 & 76.15 & 9.97 	& 2.31	\\
                MC-Net+	\cite{wu2022mutual}					(MedIA'22)	& 								& 	 							& ${83.23_{\pm1.41}}$     				& ${71.70_{\pm1.99}}$					& \textcolor{blue}{$14.92_{\pm2.56}$}	& ${3.43_{\pm0.64}}$					& -		& - 	& - 	& -		\\
                ours													& 								& 	 							& \textcolor{red}{$87.34_{\pm0.05}$}	& \textcolor{red}{$77.65_{\pm0.08}$}	& \textcolor{red}{$9.76_{\pm0.92}$}		& \textcolor{red}{$2.49_{\pm0.22}$}		& - 	& - 	& - 	& -		\\
				\hline
				UA-MT	\cite{yu2018pu}						(MICCAI'19)	& \multirow{8}{*}{$8 (10\%)$}	& \multirow{8}{*}{$72 (90\%)$}	& ${85.81_{\pm0.17}}$   				& ${75.41_{\pm0.22}}$					& ${18.25_{\pm1.04}}$					& ${5.04_{\pm0.24}}$					& 84.25 & 73.48 & 3.36 	& 13.84	\\
				SASSNet	\cite{li2020shape}					(MICCAI'20)	& 								& 	 							& ${85.71_{\pm0.87}}$    				& ${75.35_{\pm1.28}}$					& ${14.74_{\pm3.14}}$					& ${4.00_{\pm0.86}}$					& 86.81 & 76.92 & 3.94 	& 12.54	\\
				DTC		\cite{luo2021semi}  				(AAAI'21)	&								& 	  							& ${84.55_{\pm1.72}}$   				& ${73.91_{\pm2.36}}$					& ${13.80_{\pm0.16}}$					& ${3.69_{\pm0.25}}$					& - 	& - 	& - 	& -  	\\
                MC-Net	\cite{wu2021semi}					(MedIA'21)	& 								&   							& ${86.87_{\pm1.74}}$   				& \textcolor{blue}{$78.49_{\pm1.06}$}	& ${11.17_{\pm1.40}}$					& ${2.18_{\pm0.14}}$					& 87.71 & 78.31 & 9.36 	& 2.18  \\
                URPC	\cite{media2022urpc}				(MedIA'22)	& 								&   							& ${83.37_{\pm0.21}}$   				& ${71.99_{\pm0.31}}$					& ${17.91_{\pm0.73}}$					& ${4.41_{\pm0.17}}$					& - 	& - 	& - 	& -  	\\
                SS-Net	\cite{wu2022exploring}	            (MICCAI'22)	& 								&   							& ${86.56_{\pm0.69}}$   				& ${76.61_{\pm1.03}}$					& ${12.76_{\pm0.58}}$					& ${3.02_{\pm0.19}}$					& 88.55 & 79.63 & 7.49 	& 1.90  \\
                MC-Net+	\cite{wu2022mutual}					(MedIA'22)	& 								&   							& \textcolor{blue}{$87.68_{\pm0.56}$}	& ${78.27_{\pm0.83}}$					& \textcolor{blue}{$10.35_{\pm0.77}$}	& \textcolor{red}{$1.85_{\pm0.01}$}		& 88.96 & 80.25 & 7.93 	& 1.86  \\
                ours                         							& 								&   							& \textcolor{red}{$89.62_{\pm0.20}$}	& \textcolor{red}{$81.28_{\pm0.32}$}	& \textcolor{red}{$8.76_{\pm1.39}$}		& \textcolor{blue}{$2.02_{\pm0.17}$}	& - 	& - 	& - 	& -  	\\
                
				\hline
				UA-MT	\cite{yu2018pu}						(MICCAI'19)	& \multirow{8}{*}{$16 (20\%)$}	& \multirow{8}{*}{$64 (80\%)$}	& ${88.18_{\pm0.69}}$					& ${79.09_{\pm1.05}}$					& ${9.66_{\pm2.99}}$					& ${2.62_{\pm0.59}}$					& 88.88 & 80.21 & 2.26 	& 7.32  \\
				SASSNet	\cite{li2020shape}					(MICCAI'20)	& 								& 	 							& ${88.11_{\pm0.34}}$					& ${79.08_{\pm0.48}}$					& ${12.31_{\pm4.14}}$					& ${3.27_{\pm0.96}}$					& 89.27 & 80.82 & 3.13 	& 8.83  \\
				DTC		\cite{luo2021semi}  				(AAAI'21)	& 								& 	  							& ${87.79_{\pm0.50}}$					& ${78.52_{\pm0.73}}$					& ${10.29_{\pm1.52}}$					& ${2.50_{\pm0.65}}$					& 89.42 & 80.98 & 2.10 	& 7.32  \\
                MC-Net	\cite{wu2021semi}					(MedIA'21)	& 								&	  							& ${90.43_{\pm0.52}}$					& ${82.69_{\pm0.75}}$					& ${6.52_{\pm0.66}}$					& \textcolor{blue}{$1.66_{\pm0.14}$}	& 90.34 & 82.48 & 6.00 	& 1.77  \\
                URPC	\cite{media2022urpc}				(MedIA'22)	& 								&	  							& ${87.68_{\pm0.36}}$					& ${78.36_{\pm0.53}}$					& ${14.39_{\pm0.54}}$					& ${3.52_{\pm0.17}}$					& - 	& - 	& - 	& -  	\\
                SS-Net	\cite{wu2022exploring}	            (MICCAI'22)	& 								&   							& ${88.19_{\pm0.42}}$					& ${79.21_{\pm0.63}}$					& ${8.12_{\pm0.34}}$					& ${2.20_{\pm0.12}}$					& - 	& - 	& - 	& -  	\\
                MC-Net+	\cite{wu2022mutual}					(MedIA'22)	& 								&   							& \textcolor{blue}{$90.60_{\pm0.39}$}	&\textcolor{blue}{$82.93_{\pm0.64}$}	& \textcolor{blue}{$6.27_{\pm0.25}$}	& \textcolor{red}{$1.58_{\pm0.07}$}		& 91.07 & 83.67 & 5.84 	& 1.67  \\
                ours	                								& 								&   							& \textcolor{red}{$90.78_{\pm0.11}$}	& \textcolor{red}{$83.19_{\pm0.18}$}	& \textcolor{red}{$6.11_{\pm0.39}$}		& ${1.68_{\pm0.15}}$					& - 	& - 	& - 	& -  	\\
				\bottomrule[2pt] %
		      \end{tabular}
            }
		\label{tab:tab1} %
	\end{table}%

        \noindent{\textbf{Quantitative Evaluation and Comparison.}}
    	Our CAML is evaluated on four metrics: 
    	Dice, Jaccard, 
    	95$\%$ Hausdorff Distance (95HD), 
    	and Average Surface Distance (ASD). 
    	It is worth noting that the previous researchers reported results (\textbf{Reported Metrics} in Table~\ref{tab:tab1}) on LA can be confusing, 
    	with some studies reporting results from the final training iteration, 
    	while others report the best performance obtained during training. 
    	However, 
    	the latter approach can lead to overfitting of the test dataset and unreliable model selection. 
    	To ensure a fair comparison, 
    	we perform all experiments three times with a fixed set of randomly selected seeds on the same machine, 
    	and report the mean and standard deviation of the results from the final iteration.
    
    	The results on LA are presented in Table \ref{tab:tab1}. 
    	The results of the full-supervised V-Net model trained on different ratios serve as the lower and upper bounds of each ratio setting. 
    	We report the reproduced results of state-of-the-art semi-supervised methods and corresponding reported results if available. 
    	By comparing the reproduced and reported results, 
    	we observe that although the performance of current methods generally shows an increasing trend with the development of algorithms, 
    	the performance of individual experiments can be unstable.
    	and the reported results may not fully reflect the true performance.
    
        \begin{figure}[!t]
            \includegraphics[width=\textwidth]{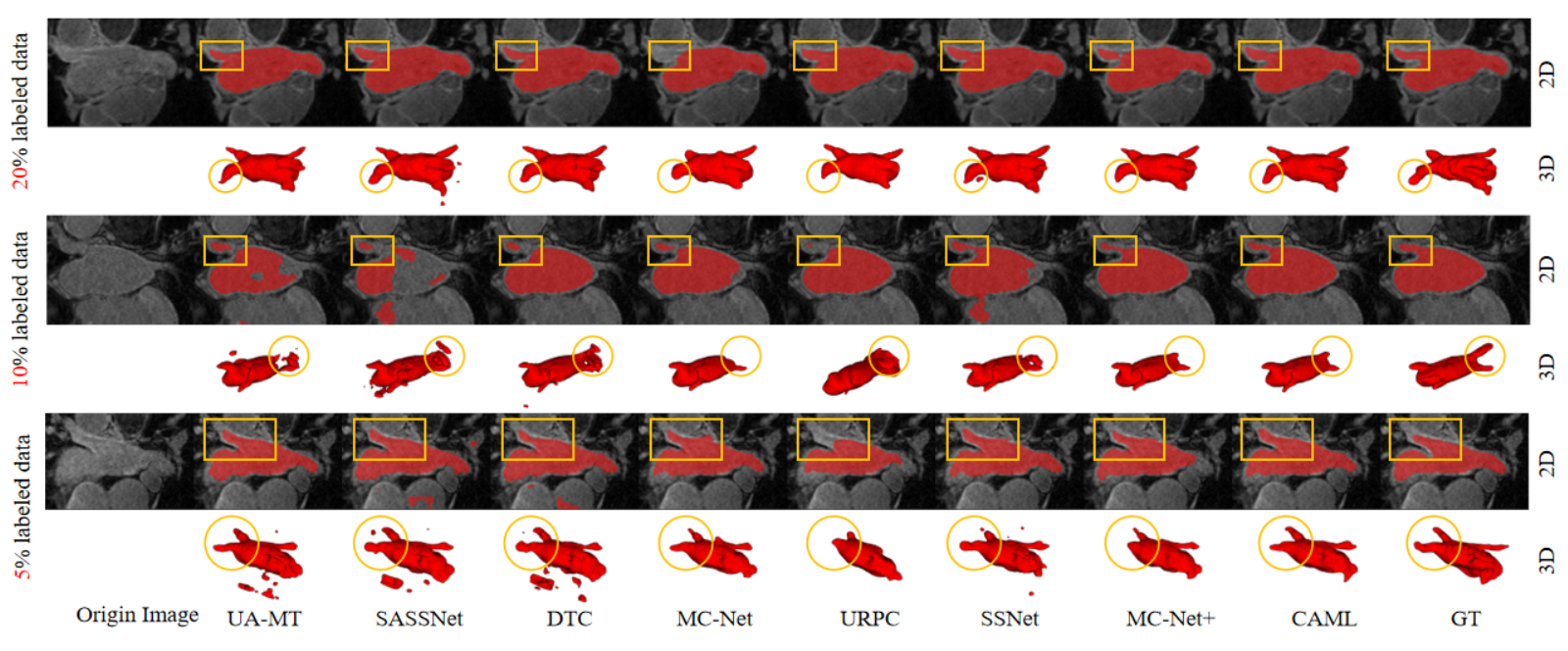}
            \caption{Visualization of the segmentations results from different methods.} 
            \label{fig:fig2}
        \end{figure}
    
    	It is evident from Table \ref{tab:tab1} that \emph{CAML} outperforms other methods by a significant margin across all settings without incurring any additional inference or post-processing costs. 
        With only $5\%$ labeled data, \emph{CAML} achieves $87.34\%$ Dice score with an absolute improvement of $4.01\%$ over the state-of-the-art. 
        \emph{CAML} also achieves $89.62\%$ Dice score with only $10\%$ labeled data.
    	When the amount of labeled data is increased to $20\%$, 
    	the model obtains comparable results with the results of V-Net trained in $100\%$ labeled data), 
    	achieving a Dice score of $90.78\%$ compared to the upper-bound model's score of $90.98\%$.
    	As presented in Table \ref{tab:tab1}, 
        through the effective transfer of knowledge between labeled and unlabeled data, 
    	\emph{CAML} achieves impressive improvements.

    	\begin{table}[!htbp]
    		\centering
                \caption{Ablation study of our proposed \emph{CAML} on the LA database.}
    		\resizebox{\textwidth}{!}{ %
    			\begin{tabular}{cc|ccc|cccc}
    				\toprule[2pt]
    				\multicolumn{2}{c|}{\textbf{Scans used}}				&
    				\multicolumn{3}{c|}{\textbf{Components}}				&
    				\multicolumn{4}{c}{\textbf{Metrics}}			\\
    				\hline
    				\makebox[0.15\textwidth][c]{Labeled} 		& 
    				\makebox[0.15\textwidth][c]{Unlabeled} 		& 
    				\makebox[0.20\textwidth][c]{Baseline} 		& 
    				\makebox[0.20\textwidth][c]{OCC} 	& 
    				\makebox[0.20\textwidth][c]{CMA}	& 
    				\makebox[0.20\textwidth][c]{Dice(\%)} 		& 
    				\makebox[0.20\textwidth][c]{Jaccard(\%)}	&
    				\makebox[0.20\textwidth][c]{95HD(voxel)}		&
    				\makebox[0.20\textwidth][c]{ASD(voxel)}	\\
    				\hline
    				\multirow{4}{*}{$4 (5\%)$}		& \multirow{4}{*}{$76 (95\%)$}	& $\surd$		& 			& 			& ${80.92_{\pm3.88}}$ 					& ${68.90_{\pm5.09}}$					& ${17.25_{\pm6.08}}$					& ${2.76_{\pm0.49}}$					\\
    												& 								& $\surd$		& $\surd$	& 			& ${83.12_{\pm2.12}}$					& ${71.73_{\pm3.04}}$					& ${16.94_{\pm7.25}}$					& ${4.51_{\pm2.16}}$	 				\\
    												& 								& $\surd$		& 			& $\surd$	& ${86.35_{\pm0.26}}$					& ${76.16_{\pm0.40}}$					& ${12.36_{\pm0.20}}$					& ${2.94_{\pm0.21}}$	 				\\
    												& 								& $\surd$		& $\surd$	& $\surd$	& \textcolor{red}{${87.34_{\pm0.05}}$}	& \textcolor{red}{${77.65_{\pm0.08}}$}	& \textcolor{red}{${9.76_{\pm0.92}}$}	& \textcolor{red}{${2.49_{\pm0.22}}$}	\\
    				\hline
    				\multirow{4}{*}{$8 (10\%)$}		& \multirow{4}{*}{$72 (90\%)$}	& $\surd$		& 			& 			& ${86.87_{\pm1.74}}$ 					& ${78.49_{\pm1.06}}$					& ${11.17_{\pm1.40}}$					& ${2.18_{\pm0.14}}$					\\
    												& 								& $\surd$		& $\surd$	& 			& ${88.50_{\pm3.25}}$					& ${79.53_{\pm0.51}}$					& ${9.89_{\pm0.83}}$					& ${2.35_{\pm0.21}}$	 				\\
    												& 								& $\surd$		& 			& $\surd$	& ${88.84_{\pm0.55}}$					& ${80.05_{\pm0.85}}$					& ${8.50_{\pm0.66}}$					& ${1.97_{\pm0.02}}$	 				\\
    												& 								& $\surd$		& $\surd$	& $\surd$	& \textcolor{red}{${89.62_{\pm0.20}}$}	& \textcolor{red}{${81.28_{\pm0.32}}$}	& \textcolor{red}{${8.76_{\pm1.39}}$}	& \textcolor{red}{${2.02_{\pm0.17}}$}	\\
    				\hline
    				\multirow{4}{*}{$16 (20\%)$}	& \multirow{4}{*}{$64 (80\%)$}	& $\surd$		& 			& 			& ${90.43_{\pm0.52}}$ 					& ${82.69_{\pm0.75}}$					& ${6.52_{\pm0.66}}$					& \textcolor{red}{${1.66_{\pm0.14}}$}	\\
    												& 								& $\surd$		& $\surd$	& 			& ${90.27_{\pm0.22}}$					& ${82.42_{\pm0.39}}$					& ${6.96_{\pm1.03}}$					& ${1.91_{\pm0.24}}$	 				\\
    												& 								& $\surd$		& 			& $\surd$	& ${90.25_{\pm0.28}}$					& ${82.34_{\pm0.43}}$					& ${6.95_{\pm0.09}}$					& ${1.79_{\pm0.18}}$	 				\\
    												& 								& $\surd$		& $\surd$	& $\surd$	& \textcolor{red}{${90.78_{\pm0.11}}$}	& \textcolor{red}{${83.19_{\pm0.18}}$}	& \textcolor{red}{${6.11_{\pm0.39}}$}	& ${1.68_{\pm0.15}}$					\\
    				\bottomrule[2pt]
    			\end{tabular}
    		} %
    		\label{tab:tab2} %
    	\end{table}%
    
        Table \ref{tab:tab1} also demonstrated that as the labeled data ratio declines, 
            the model maintains a low standard deviation of results, 
            which is significantly lower than other state-of-the-art methods.
    	This finding suggests that \emph{CAML} is highly stable and robust. 
    	Furthermore, the margin between our method and the state-of-the-art semi-supervised methods increases with the decline of the labeled data ratio, 
    	indicating that our method rather effectively transfers knowledge from labeled data to unlabeled data, thus enabling the model to extract more universal features from unlabeled data.
    	Figure \ref{fig:fig2} shows the qualitative comparison results. 
    	The figure presents 2D and 3D visualizations of all the compared methods
    	and the corresponding ground truth. 
            As respectively indicated by the orange rectangle and circle in the 2D and 3D visualizations
    	Our CAML achieves the best segmentation results compared to all other methods. 

        \subsubsection{Ablation Study.}
            In this section, we analyze the effectiveness of the proposed \emph{CMA} module and \emph{OCC} module. 
            We implement the MC-Net as our baseline, which uses different up-sampling operations to introduce architecture heterogeneity.
            Table \ref{tab:tab2} presents the results of our ablation study.
        The results demonstrate that under $5\%$ ratio, 
    	both \emph{CMA} and \emph{OCC} significantly improve the performance of the baseline. 
    	By combining these two modules, 
    	\emph{CAML} achieves an absolute improvement of $6.42\%$ in the Dice coefficient.
            Similar improvements can be observed for a data ratio of $10\%$.
    	Under a labeled data ratio of $20\%$, 
    	the baseline performance is improved to $90.43\%$ in the Dice coefficient, 
    	which is approximately comparable to the upper bound of a fully-supervised model. 
    	In this setting, 
    	adding the \emph{CMA} and \emph{OCC} separately may not achieve a significant improvement. 
    	Nonetheless, by combining these two modules in our proposed \emph{CAML} framework, 
    	we still achieve the best performance in this setting,
    	which further approaches the performance of a fully-supervised model.

    \section{Conclusion}
            In this paper,
            we proposed a novel framework named \emph{CAML} for semi-supervised medical image segmentation. Our key idea is that cross-sample correlation should be taken into consideration for semi-supervised learning. To this end, two novel modules: \emph{Cross-sample Mutual Attention(CMA)} and \emph{Omni-Correlation Consistency(OCC)} are proposed to encourage efficient and direct transfer of the prior knowledge from labeled data to unlabeled data. 
            Extensive experimental results on the LA dataset demonstrate that we outperform previous state-of-the-art results by a large margin without extra computational consumption in inference.
    \\

    \noindent\textbf{Acknowledgements.}
    This work is funded by the Scientific and Technological Innovation 2030 New Generation Artificial Intelligence Project of the National Key Research and Development Program of China (No.2021ZD0113302), 
    Beijing Municipal Science and Technology Planning Project (No.Z201100005620008, Z211100003521009).

    \bibliography{reference}
	
\end{document}